\documentclass[10pt,twocolumn,letterpaper]{article}
\usepackage{tikz} 
\usepackage{pgfplots} 
\usepackage{times} 
\usepackage{epsfig} 
\usepackage{graphicx} 
\usepackage{amsmath} 
\usepackage{amssymb} 
\usepackage{amsthm}
\usepackage{booktabs} 
\pgfplotsset{compat=newest}
\pgfplotsset{plot coordinates/math parser=false}
\newlength\figureheight
\newlength\figurewidth 
\usepackage[utf8]{inputenc}

\begin{document}
\title{A fast minimal solver for absolute camera pose with unknown focal length and radial distortion from four planar points} 
\author{Magnus Oskarsson\\Centre for Mathematical Sciences\\Lund University\\Lund, Sweden\\{\tt\small magnuso@maths.lth.se}
}
\maketitle 
\begin{abstract} 
In this paper we present a fast minimal solver for absolute camera pose estimation from four known points that lie in a plane. We assume a perspective camera model with unknown focal length and unknown radial distortion. The radial distortion is modelled using the division model with one parameter. We show that the solutions to this problem can be found from a univariate six-degree polynomial. This results in a very fast and numerically stable solver.    
\end{abstract}
\section{Introduction}
The camera pose estimation problem is a classic problem in computer vision. Given a number of known 3D-points, and their projections in a camera, the object is to estimate the pose of the camera, i.e.\ the location and viewing direction. This geometric problem has been studied for a very long time. As an example a solution was given by Grunert already in 1841 \cite{grunert1841pothenotische}. For a fully calibrated camera the minimal number of points is three. Different number of 3D-points and different knowledge of the calibration of the camera will give different problems. 
  We will in this paper give a solution to one such special pose problem, namely \emph{camera pose estimation from four coplanar 3D-points, where the unknown radial distortion and unknown focal length of the camera are also estimated.} This was one of the problems studied and solved in \cite{bujnak2010new}. For the more general problem, where the points need not lie in a plane, the state-of-the-art solver in terms of speed was presented in \cite{larsson2017making}. The same solver can also be used for the planar case. However, in this paper we will present a simpler solution, that gives a faster solver.  

\section{Parametrization of the problem}
We will use the same parametrization that was described in \cite{bujnak2010new}, and we include the derivations here for completeness. 
We have four given 3D-points, that all lie in a plane. We can without loss generality assume that these lie in the plane $Z=0$. This means that we represent our 3D-points in homogeneous coordinates as 
\begin{equation} 
U_i = \begin{bmatrix} X_i \\Y_i \\1 \end{bmatrix},\quad i = 1,\ldots,4.
\end{equation}
Since we have dropped the third coordinate of our points, we simply remove the third column in the camera matrix in our projection equation. We have a calibrated camera with unknown focal length, $P = KP_n$, i.e.\ 
\begin{equation}
P =  \begin{bmatrix}1 & 0 & 0\\0 & 1 & 0\\0 & 0 & w\end{bmatrix}
\begin{bmatrix}r_{11} & r_{12} & t_x \\r_{21} & r_{22} & t_y \\r_{31} & r_{32} & t_z \end{bmatrix},
\end{equation}
where $w = 1/f$.
Using the division model for radial distortion \cite{fitzgibbon2001simultaneous}, our undistorted image coordinate can be represented in homogeneous coordinates as 
\begin{equation}
u_i = \begin{bmatrix} x_i\\y_i\\1+kr_i^2\end{bmatrix}, \quad  i=1,\ldots,4, 
\end{equation}
where $r_i^2 = x_i^2+y_i^2$.
This gives us the following projection equations 
\begin{equation}
\lambda_i u_i = PU_i, \quad i = 1,\ldots,4.
\end{equation}
Taking the cross-product with with $u_i$ eliminates the unknown depths $\lambda_i$,
\begin{equation}
 u_i \times PU_i = {\bf 0},\quad i = 1,\ldots,4.
\label{e_constraints1}
\end{equation}
This gives for each point three constraints, out of which only two are linearly independent. The third such constraint for each point is special in the sense that it neither involves the unknown radial distortion  parameter $k$ nor the focal length  parameter $w$. If we take these constraints using all four points we get four linear constraints  on the six elements of the first two rows of $P$. Since the scale of the camera is arbitrary we can write the first two rows of $P$ using a single unknown 
$\beta$.
The second row of (\ref{e_constraints1}) is 
\begin{equation}
(1+kr_i^2)(p_{11}X_i+p_{12}Y_i+p_{14})-x_i(p_{31}X_i+p_{32}Y_i+p_{34})=0
\end{equation}
The elements from the first row of $P$ are already parametrized with $\beta$. In addition we have elements from the third row of $P$. We now use three of the constraints in (\ref{e_constraints1}) and write them as 
\begin{equation}
C_{3\times 3} \begin{bmatrix}p_{31}\\p_{32}\\p_{34}\end{bmatrix} = D_{3\times 4}\begin{bmatrix}\beta \\ k\beta\\k\\1\end{bmatrix},
\end{equation}
where $C$ and $D$ only depend on known data.
If $U_1$, $U_2$ and $U_3$ are not collinear we can express the elements in $P$ as 
\begin{equation}
 \begin{bmatrix}p_{31}\\p_{32}\\p_{34}\end{bmatrix} = C^{-1}D \begin{bmatrix}\beta \\ k\beta\\k\\1\end{bmatrix}.
\end{equation}
The first two columns of $P_n$ are from scaled rotation matrix, and they must hence be orthogonal and of equal length. We can express these two constraints using the elements of $P$ by introducing the two vectors $v_1 = [ wp_{11} \,\, wp_{21} \,\, p_{31} ]^T$ and  $v_2 = [ wp_{12} \,\, wp_{22} \,\, p_{32} ]^T$ as 
\begin{align}
f_1 &= v_1^Tv_2= 0,\\
f_2 &= v_1^Tv_1 - v_2^Tv_2 = 0.
\end{align} 
These two equations together with the fourth constraint (which we denote $f_3=0$) from (\ref{e_constraints1}) give us the three equations needed to solve for the unknown $w$, $\beta$ and $k$. In \cite{bujnak2010new} they solved this system using the action matrix method. In the next section we will show how we can instead find a univariate polynomial in $\beta$.
 
\section{Our new solver}
The final constraint $f_3=0$ only involves the unknown $k$ and $\beta$ and can be written 
\begin{equation} 
q_{31}k\beta+q_{32}k+q_{33}\beta + q_{34}=0,
\end{equation}
where $q_{ij}$ only depend on data.
If we solve for $k$ we can write the solution as
\begin{equation}
k = -\frac{q_{33}\beta+q_{34}}{q_{31}\beta+q_{32}}.
\label{e_ksol}
\end{equation}
Inserting the solution into $f_1=0$ and $f_2= 0$ gives two new equations in only $\beta$ and $w$. These are rational functions, but multiplying by an appropriate power of the denominator in (\ref{e_ksol}) gives two new polynomial constraints $f'_1=0$ and $f'_2=0$. The original polynomials $f_1$ and $f_2$ involved only constant factors and powers of twos in terms of the unknown $w$. Since the solution in (\ref{e_ksol}) doesn't involve $w$ the same will hold true for $f'_1$ and $f'_2$. This means that they can be written in the following form
\begin{align}
q_{11}(\beta)w^2+q_{12}(\beta) &= 0,\\
q_{21}(\beta)w^2+q_{22}(\beta) &= 0.
\end{align} 
These equations can be written in matrix form as
\begin{equation}
BW = 0,
\label{e_bw}
\end{equation}
with $W = [w^2\,\, 1]^T$ and
\begin{equation}
B = \begin{bmatrix}q_{11}(\beta) & q_{12}(\beta) \\
q_{21}(\beta) & q_{22}(\beta) \end{bmatrix}.
\end{equation}
Since (\ref{e_bw}) should have a non-trivial solution then $det(B)=0$. This gives us our sought univariate polynomial in $\beta$. The determinant constructed in this way contains the denominator of (\ref{e_ksol}) as a factor but this can easily be factored out symbolically. This leaves a six-degree polynomial whose coefficients can be determnined directly from the input data. This gives six potential solutions for $\beta$. The corresponding solutions for $w$ are then found from the nullspace of $B$ and the solutions for $k$ are found from (\ref{e_ksol}). 
The entries of the camera matrix $p_{ij}$ are parametrized in $\beta$ and $k$, and this gives us the corresponding solutions for the translation and the first two colums of the rotation matrix. The last column is obtained by the cross-product of the first two. Note that if, for a given solution $p_{ij}$, all the depths $\lambda$ are negative we should choose $-p_{ij}$ as the solution (If the depths have different signs or are complex we disregard this solution, since it is not a physical solution).

\section{Evaluation of solver}
\begin{figure}
\centering
\resizebox{0.45\textwidth}{!}{
%
%
\definecolor{mycolor1}{rgb}{0.00000,0.44700,0.74100}%
\begin{tikzpicture}

\begin{axis}[%
width=4.553in,
height=3.262in,
at={(0.764in,0.495in)},
axis x line=bottom,
axis y line=left, 
scale only axis,
xmin=-20,
xmax=-3,
xtick={-20,-18,...,-3},
xlabel style={font=\color{white!15!black}},
xlabel={$\text{log}_{\text{10}}\text{ residuals}$},
ymin=0,
ymax=0.45,
ytick={0,0.1,...,0.5},
ylabel style={font=\color{white!15!black}},
ylabel={\%},
axis background/.style={fill=white}
]
\addplot [color=mycolor1, line width=3.0pt, forget plot]
  table[row sep=crcr]{%
-20	0\\
-19.8	0\\
-19.6	0\\
-19.4	0\\
-19.2	0\\
-19	0.000172529804523731\\
-18.8	8.62649022618657e-05\\
-18.6	0.000258794706785597\\
-18.4	0.000776384120356792\\
-18.2	0.000517589413571194\\
-18	0.00060385431583306\\
-17.8	0.000776384120356792\\
-17.6	0.00189782784976105\\
-17.4	0.00293300667690343\\
-17.2	0.00465830472214075\\
-17	0.00741878159452045\\
-16.8	0.0113869670985663\\
-16.6	0.0196683977157054\\
-16.4	0.0328669277617708\\
-16.2	0.0515864115525957\\
-16	0.0855747830437708\\
-15.8	0.127585790445299\\
-15.6	0.18124255965218\\
-15.4	0.241282931626438\\
-15.2	0.296061144562723\\
-15	0.337727092355204\\
-14.8	0.348251410431152\\
-14.6	0.3467849070927\\
-14.4	0.331688549196874\\
-14.2	0.320646641707355\\
-14	0.289159952381774\\
-13.8	0.263970600921309\\
-13.6	0.24180052104001\\
-13.4	0.220148030572281\\
-13.2	0.186677248494677\\
-13	0.16459343351564\\
-12.8	0.138455168130295\\
-12.6	0.120080743948517\\
-12.4	0.0999810217215024\\
-12.2	0.0799675643967495\\
-12	0.0735839616293715\\
-11.8	0.0632321733579476\\
-11.6	0.0512413519435482\\
-11.4	0.043046186228671\\
-11.2	0.0349372854160556\\
-11	0.0296751263780818\\
-10.8	0.0246717620468936\\
-10.6	0.0213936957609427\\
-10.4	0.0194958679111817\\
-10.2	0.0152688877003502\\
-10	0.0125946757302324\\
-9.8	0.00974793395559083\\
-9.6	0.00776384120356792\\
-9.4	0.00681492727868739\\
-9.2	0.00526215903797381\\
-9	0.00353686099273649\\
-8.8	0.00508962923345008\\
-8.6	0.0032780662859509\\
-8.4	0.00198409275202291\\
-8.2	0.00215662255654664\\
-8	0.000948913924880523\\
-7.8	0.000948913924880523\\
-7.6	0.00129397353392799\\
-7.4	0.00060385431583306\\
-7.2	0.000517589413571194\\
-7	0.00060385431583306\\
-6.8	0.000517589413571194\\
-6.6	0.000345059609047463\\
-6.4	0.000517589413571194\\
-6.2	8.62649022618657e-05\\
-6	8.62649022618657e-05\\
-5.8	0.000431324511309329\\
-5.6	8.62649022618657e-05\\
-5.4	8.62649022618657e-05\\
-5.2	8.62649022618657e-05\\
-5	0.000258794706785597\\
-4.8	8.62649022618657e-05\\
-4.6	0\\
-4.4	8.62649022618657e-05\\
-4.2	0.000258794706785597\\
-4	0\\
-3.8	0\\
};
\end{axis}
\end{tikzpicture}%
}
\caption{Error histogram for 100,000 random instances of the four-point pose problem,
 on a log-scale.}
\label{f_numerics}
\end{figure}
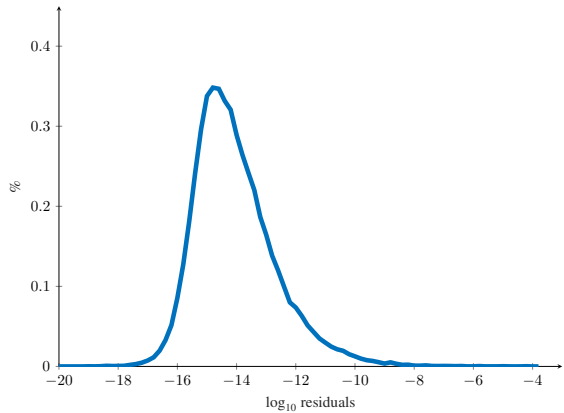

We have implemented a version of this solver in Matlab.\footnote{ The code is publicly available at \\ \emph{https://github.com/hamburgerlady/fast\_planar\_camera\_pose}}
The complete solver including all datamanipulation and extraction of rotation matrices runs in under  0.3 ms on a standard desktop computer.
In order to test the numerical stability of our solver we generated 100,000 random instances of four corresponding image and world points. We then ran our solver and extracted the solutions for each instance. Using the estimated camera pose and calibration parameters we can calculate the reprojected image points and compare with the input image points. A histogram of the reprojection errors on a log-scale can be seen in Figure~\ref{f_numerics}. As we are estimating our parameters from a minimal set of datapoints, in theory we should have zero reprojection error. Due to numerical instabilities in solvers this may not be the case, but as can be seen from the graph our solver performs close to machine precision.
 \section{Conclusion}
We have presented some simple derivations that give the means for  formulating the camera pose problem with unknown radial distortion and unknown focal length given four coplanar points in terms of  a univariate six-degree polynomial. By finding the roots of this polynomial we can find the six possible solutions to the pose problem. This gives a stable and very fast solver.    
{\small
\bibliographystyle{ieee}
\bibliography{mainbib}

\begin{thebibliography}{1}\itemsep=-1pt

\bibitem{bujnak2010new}
M.~Bujnak, Z.~Kukelova, and T.~Pajdla.
\newblock New efficient solution to the absolute pose problem for camera with
  unknown focal length and radial distortion.
\newblock In {\em Asian Conference on Computer Vision}, pages 11--24. Springer,
  2010.

\bibitem{fitzgibbon2001simultaneous}
A.~W. Fitzgibbon.
\newblock Simultaneous linear estimation of multiple view geometry and lens
  distortion.
\newblock In {\em Computer Vision and Pattern Recognition, 2001. CVPR 2001.
  Proceedings of the 2001 IEEE Computer Society Conference on}, volume~1, pages
  I--I. IEEE, 2001.

\bibitem{grunert1841pothenotische}
J.~A. Grunert.
\newblock Das pothenot'sche {P}roblem in erweiterter {G}estalt; nebst
  bemerkungen \"{u}ber seine {A}nwendung in der {G}eodasie.
\newblock {\em Grunert Archiv der Mathematik und Physik}, pages 238--248, 1841.

\bibitem{larsson2017making}
V.~Larsson, Z.~Kukelova, and Y.~Zheng.
\newblock Making minimal solvers for absolute pose estimation compact and
  robust.
\newblock In {\em 2017 IEEE International Conference on Computer Vision
  (ICCV)}, pages 2335--2343. IEEE, 2017.

\end{thebibliography}
}

\end{document}